\documentclass[conference]{IEEEtran}
\IEEEoverridecommandlockouts
\usepackage{cite}
\usepackage{amsmath,amssymb,amsfonts}
\usepackage{algorithmic}
\usepackage{graphicx}
\usepackage{textcomp}
\usepackage{xcolor}
\usepackage{booktabs}
\usepackage{multirow}
\usepackage{balance}
\usepackage{url}
\usepackage[utf8]{inputenc}
\usepackage[T1]{fontenc}
\usepackage{times}
\def\BibTeX{{\rm B\kern-.05em{\sc i\kern-.025em b}\kern-.08em
    T\kern-.1667em\lower.7ex\hbox{E}\kern-.125emX}}

\usepackage{hyperref}
\hypersetup{
  colorlinks   = true, 
  urlcolor     = blue, 
  linkcolor    = blue, 
  citecolor    = blue 
}

\usepackage{caption}
\usepackage{etoolbox}
\newcommand{\insertfig}{%
  \setcounter{figure}{0}  
  \includegraphics[width=0.8\linewidth]{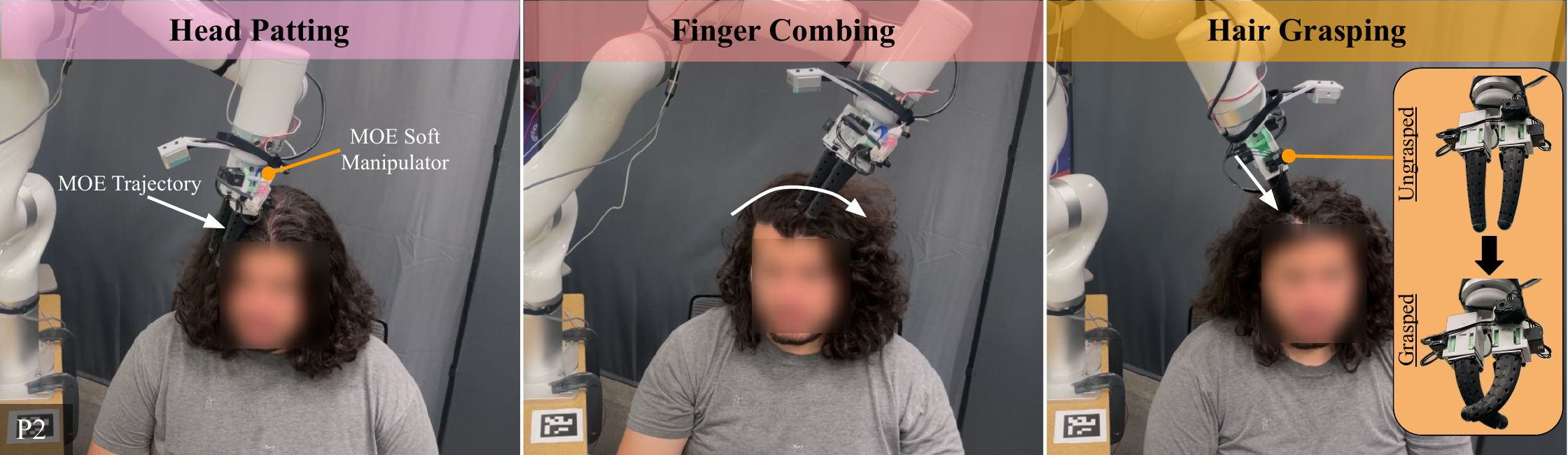}
  \captionof{figure}{\textbf{Hair care skills studied with the proposed MOE-Hair system} demonstrated with Participant 2 (P2). Left: Multi-finger Omnidirectional End-effector (MOE) contacts the subject's head to perform head patting. Middle: MOE follows a user-defined trajectory across the head to perform the finger-combing skill. Right: MOE approaches the user's scalp and grasps hair using soft dexterous fingers.  }
  \label{fig:front}
}

\makeatletter
\apptocmd{\@maketitle}{\centering\insertfig}{}{} 
\makeatother

\begin{document}

\title{Soft and Compliant Contact-Rich Hair \\Manipulation and Care}

\makeatletter
\newcommand{\linebreakand}{%
  \end{@IEEEauthorhalign}
  \hfill\mbox{}\par
  \mbox{}\hfill\begin{@IEEEauthorhalign}
}
\makeatother

\author{
  \IEEEauthorblockN{Uksang Yoo}
  \IEEEauthorblockA{\textit{Carnegie Mellon University} \\
    \texttt{uksang@cmu.edu}}
  \and
  \IEEEauthorblockN{Nathaniel Dennler}
  \IEEEauthorblockA{\textit{University of Southern California} \\
    \texttt{dennler@usc.edu}}
  \and
  \IEEEauthorblockN{Eliot Xing}
  \IEEEauthorblockA{\textit{Carnegie Mellon University} \\
    \texttt{etaoxing@cmu.edu}}
  \linebreakand 
  \IEEEauthorblockN{Maja Matarić}
  \IEEEauthorblockA{\textit{University of Southern California} \\
    \texttt{mataric@usc.edu}}
  \and
  \IEEEauthorblockN{Stefanos Nikolaidis}
  \IEEEauthorblockA{\textit{University of Southern California} \\
    \texttt{nikolaid@usc.edu}}
  \linebreakand 
  \IEEEauthorblockN{Jeffrey Ichnowski}
  \IEEEauthorblockA{\textit{Carnegie Mellon University} \\
    \texttt{jeffi@cmu.edu}}
  \and
  \IEEEauthorblockN{Jean Oh}
  \IEEEauthorblockA{\textit{Carnegie Mellon University} \\
    \texttt{jeanoh@cmu.edu}}
}


\maketitle

\begin{abstract}
Hair care robots can help address labor shortages in elderly care while enabling those with limited mobility to maintain their hair-related identity. We present MOE-Hair, a soft robot system that performs three hair-care tasks: head patting, finger combing, and hair grasping. The system features a tendon-driven soft robot end-effector (MOE) with a wrist-mounted RGBD camera, leveraging both mechanical compliance for safety and visual force sensing through deformation. In testing with a force-sensorized mannequin head, MOE achieved comparable hair-grasping effectiveness while applying significantly less force than rigid grippers. Our novel force estimation method combines visual deformation data and tendon tensions from actuators to infer applied forces, reducing sensing errors by up to 60.1\% and 20.3\% compared to actuator current load-only and depth image-only baselines, respectively. A user study with 12 participants demonstrated statistically significant preferences for MOE-Hair over a baseline system in terms of comfort, effectiveness, and appropriate force application. These results demonstrate the unique advantages of soft robots in contact-rich hair-care tasks, while highlighting the importance of precise force control despite the inherent compliance of the system. Videos, data, and code are available at~\href{https://moehair.github.io}{\texttt{moehair.github.io}}.
\end{abstract}

\begin{IEEEkeywords}
Assistive robotics, Soft robotics, Manipulation
\end{IEEEkeywords}

\section{Introduction}
Hair plays an important role in people's identities and self-esteem~\cite{batchelor_hair_2001,lashley_importance_2020, yoo2024inclusion}. Notably, the importance of hair to a person's self-esteem tends to increase with age~\cite{ward_if_2011}. With aging and loss of independent mobility, hair care becomes an increasingly time-consuming and difficult daily task that is also crucial for personal hygiene~\cite{edemekong2023activities}. Despite this, most elder care and hospice facilities heavily rely on volunteers for hair-care assistance ~\cite{burbeck_understanding_2014}. Toward addressing the gap in hair-care services, researchers proposed deploying robot assistance for combing~\cite{dennler_design_2021, hughes_detangling_2021}. A notable challenge in previous works was that human subjects tend to perceive rigid robots as being ``rough''~\cite{dennler_design_2021}. Hair care and manipulation tasks additionally pose a perception challenge for robot systems because hair can often occlude the underlying scalp, making consistent application of force on the head difficult from just external vision. Previous approaches with mechanically rigid robot manipulators used high-cost force sensors to interact with the head safely~\cite{hughes_detangling_2021}.

Soft robot manipulators have many advantages in addressing challenges in close contact with human users. Their compliance makes them safer in unstructured environments~\cite{low_sensorized_2022} and more robust in contact-rich manipulation tasks~\cite{bhatt_surprisingly_2021}. As such, soft robot manipulators are particularly useful in physical human-robot interaction (pHRI) tasks~\cite{polygerinos_soft_2017, yoo2024moe}. Human subjects tend to perceive soft robots as safer than their rigid counterparts~\cite{jorgensen2022soft}. Additionally, soft robot manipulators deform when they make contact, in contrast to rigid robots~\cite{yoo_toward_2023}. Observing these deformations offers a promising direction for using soft robots for interactive perception~\cite{bohg2017interactive}.

\begin{figure}[t!]
\centering
\includegraphics[width=0.7\columnwidth]{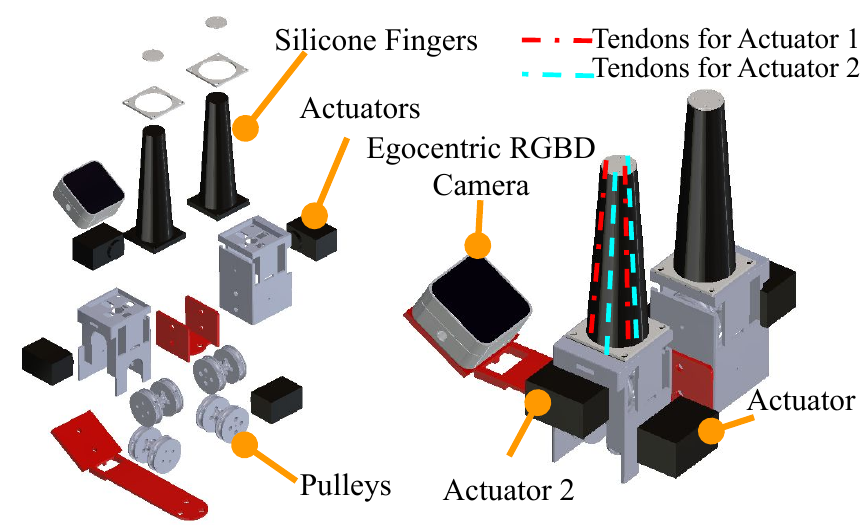}
  \caption{\textbf{Design of proposed MOE end-effector.} Left: exploded view and assembly of MOE. Right: fully assembled MOE.}
  \label{figure:design}
  \vspace{-0.5 cm}
\end{figure}

In this work, we propose a dexterous tendon-driven soft robot manipulator that we call Multi-fingered Omnidirectional End-effector (MOE) for hair-care applications (Fig. \ref{fig:front}). We first validate the physical safety of MOE using a force-sensorized mannequin head and then validate the efficacy of MOE for hair-care tasks in a user study. We demonstrate with a testbed that MOE's compliant fingers make them appropriate for human contact. We show that MOE applies less force on the head compared to a rigid gripper, given the same hair depth, while being equally effective in grasping hair. Despite MOE's compliance, the ability to reason about contact and applied forces is still important to maintaining contact with the user's head and to apply consistent forces. To this end, we propose a method to infer MOE's applied forces on the head by observing its deformation and the tendon tensions that are indirectly observed with the actuator current load. Finally, we present MOE-Hair, a system that incorporates the proposed MOE hand and the applied force estimation method to perform three hair care skills: head patting, finger combing, and hair grasping. We find that users interacting with MOE rate it as effective, comfortable, and appropriately forceful across the three hair care skills we evaluated.

To summarize, we make the following contributions:
\begin{itemize}
    \item Design of a novel dexterous tendon-driven soft robot manipulator that we call MOE for hair care tasks, 
    \item Method for indirectly estimating contact force direction and magnitude from vision and tendon tensions measured from the actuators, 
    \item MOE-Hair, a system that uses proposed MOE and force sensing methods to perform hair-care tasks,
    \item Evaluation of MOE-Hair on a force-sensorized mannequin head and with a user study, evaluating task effectiveness, user comfort, and appropriate use of force. 
\end{itemize}

\section{Related Work}

\subsection{Hair Care Robots}

Robots are demonstrating increasing capabilities to assist users in many Activities of Daily Living (ADLs)~\cite{bilyea2017robotic}. Researchers studied applications such as feeding~\cite{sundaresan2022learning,jenamani2024feel,bhattacharjee2020more, gallenberger2019transfer}, bathing~\cite{madan2024rabbit,erickson2022characterizing}, and dressing~\cite{clegg2020learning,zhang2022learning}. Researchers identified that individuals perceived hair care to be important for preserving well-being~\cite{lomborg2005body} and dignity~\cite{ward_if_2011} as with other ADLs.

Because hair grooming and care tasks are labor intensive, researchers recently proposed automated and robot solutions~\cite{dennler_design_2021}. Previous works in hair-care robots focused on visually estimating hair flow and either following the existing hair flow directions~\cite{dennler_design_2021} or using a special sensorized brush and a feedback controller with a high fidelity force sensor attached to the end-effector~\cite{hughes_detangling_2021}. While these works use systems with rigid grippers, in contrast, we use soft manipulators. Additionally, this work presents the first user study results for evaluating user responses to a robot manipulator touching and manipulating their hair. 

\begin{figure}[t!]
\centering
\includegraphics[width=0.8\columnwidth]{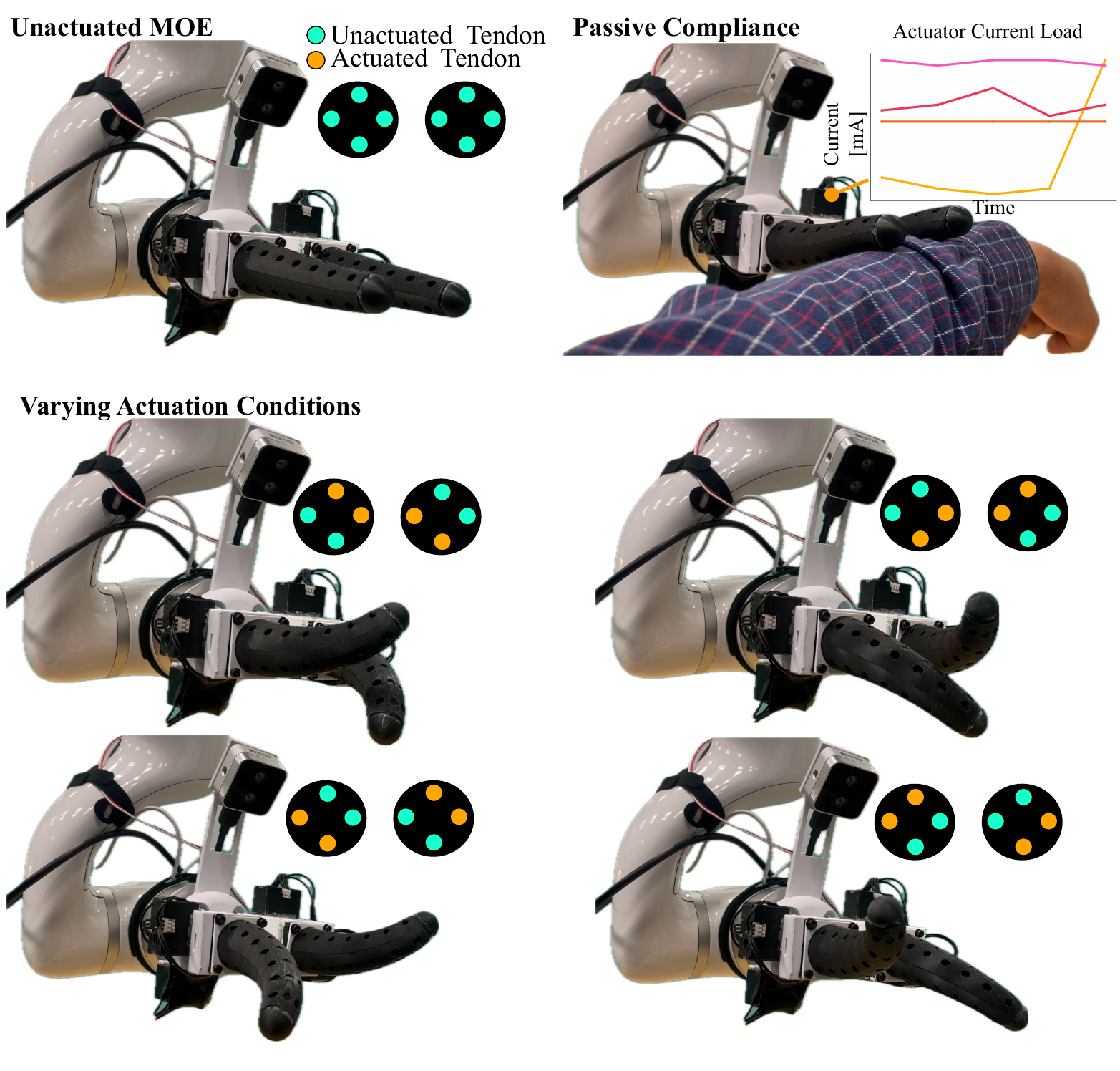}
  \caption{\textbf{MOE in varying poses.} MOE deforms easily with contact by inherent passive compliance of the fingers' material. As the fingers deform, they apply more tension on the tendons, which our proposed method uses in conjunction with depth image to estimate applied forces. The tendons can be actuated in various ways to achieve compliant dexterity of the fingers.}
  \label{figure:deformed}
  
\end{figure}

\subsection{Assistive Soft Robots}
Researchers have studied soft robot manipulators' advantages in various delicate manipulation tasks including food handling~\cite{low_sensorized_2022}, crop handling~\cite{elfferich_soft_2022}, minimally invasive surgeries~\cite{fu2021toward,kuntz2020learning,runciman_soft_2019, yoo_analytical_2021}, and wearable assistive robots~\cite{walsh_human---loop_2018}. In such applications, researchers noted that the inherent compliance of the soft robots enables reliable and safe performance~\cite{polygerinos_soft_2017}.

Importantly, soft robots have notable advantages in assistive applications, especially in settings where the robot has to make close contact with human users~\cite{polygerinos2017soft}. The deformable and compliant material of the soft robots allows them to safely interact with the human body~\cite{cheng2020brain,manti2016soft} and deform around the human body to apply distributed pressure~\cite{hedayati2019hugbot, liu2024skingrip}.

Researchers showed that users generally perceive soft robots to be friendly and safe~\cite{jorgensen2022soft, koike2024sprout}. Recent works demonstrated that compliance in soft manipulators designed for certain assistive tasks such as bathing can outperform rigid robots in perceived comfort, effectiveness, and safety~\cite{liu2024skingrip}. To our knowledge, this paper presents the first exploration of the opportunities and advantages presented uniquely by soft robot manipulators in hair care.
\begin{figure}[t!]
  \includegraphics[width=0.9\columnwidth]{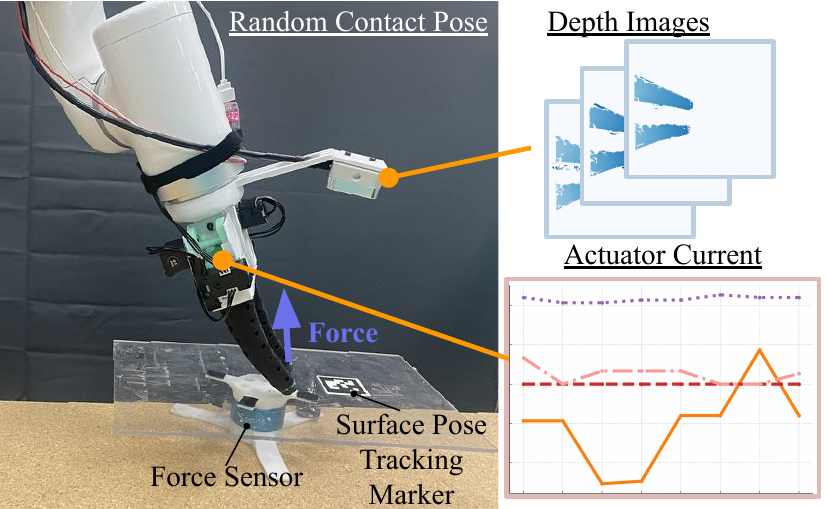}
  \caption{\textbf{Setup for training data collection}. We use a controlled setup with a force sensor and allow MOE to randomly sample various contact conditions, capturing training observations and forces in a self-labeling manner to scale up data collection.}
  \label{figure:data_collection}
\end{figure}

\subsection{Sensing for Soft Robots}
While soft robots offer many advantages in robust manipulation in uncertain environments~\cite{bhatt_surprisingly_2021} and safe interaction with delicate objects~\cite{sinatra2019ultragentle}, sensorizing them is an active challenge~\cite{wang2018toward, zhou2024integrated}. Because of the soft robots' continuously deforming structure, embedding popular rigid tactile or contact sensors such as GelSight~\cite{yuan2017gelsight} and Digit~\cite{lambeta2020digit} is difficult and would result in introducing rigidity and undercutting advantages of fully soft robots. To address these challenges, researchers have proposed various specialized soft sensors that can be embedded into the soft robot~\cite{wall2017method, tapia2020makesense,roberts2021soft}. 

To avoid using such specialized soft sensors that can be expensive and difficult to fabricate~\cite{wang2018toward} and require robots specifically designed to allow the sensor to be embedded~\cite{tapia_makesense_2020}, researchers proposed using external vision-based approaches and reasoning about contact conditions based on perceived deformation~\cite{collins2023force, grady2022visual}. We extend such methods to incorporate robot actuator loads into inferring soft robot manipulator's applied forces and use a foundation segmentation model~\cite{ravi2024sam} to train the network to only consider the soft robot's deformation. We demonstrate that such force estimation for soft robot manipulators results in a significantly improved perception of the system's effectiveness, and appropriate usage of force. By doing so, this work introduces the opportunities and advantages of soft robot manipulators for hair care.


\section{Methodology}

\subsection{MOE Soft Robot Design}
We introduce a dexterous soft tendon-driven manipulator that we call Multi-finger Omnidirectional End-effector (MOE). Fig.~\ref{figure:design} shows MOE with two soft fingers molded from low-hardness silicone (Ecoflex 00-30, Smooth-on). Two servomotors activate each finger through four embedded tendons. Each pair of tendons actuated by a single servo actuator controls the MOE finger's range of motion in a bending plane. The design is modular---each finger is an independent detachable subsystem that can be assembled in approximately 2 hours. The total component cost is 375 USD. We chose a fully soft design to maintain the human finger-like form factor (105 mm length, 17 mm in diameter) and the beneficial dexterity for hair care tasks. MOE's design can be extended to variants with more fingers as needed. For this work, we determined that two fingers are sufficient. We placed an RGBD camera (Realsense D405, Intel) on the wrist of MOE to provide egocentric view depth images.




\begin{figure}[t!]
    \centering
  \includegraphics[trim={0.6cm 0 0 0},clip,width=1.0\columnwidth]{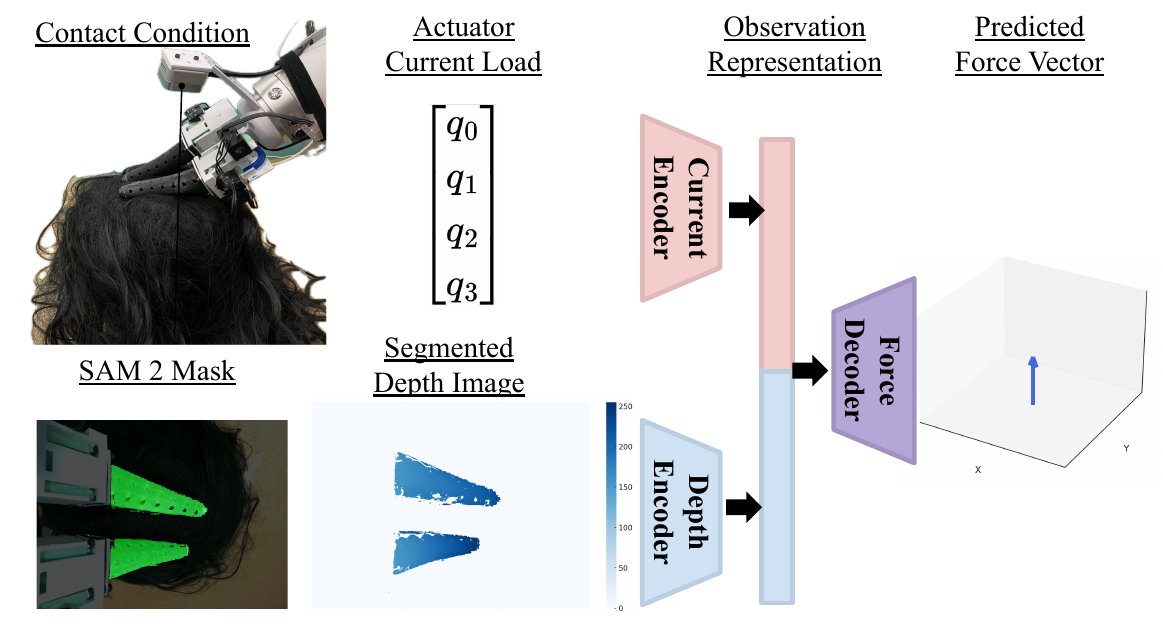}
  \caption{\textbf{Proposed method for force estimation.} The proposed applied force estimation module uses segmented depth image and actuator current loads as input to predict the force vector.  }
  \label{figure:force_module}
\end{figure}

\subsection{Force Estimation Module}

To enable MOE to maintain contact with the head during tasks and perform tasks effectively and comfortably for the users, we developed methods for predicting MOE contact forces. Using the wrist-mounted egocentric RGBD camera, we can capture depth images of MOE as it deforms with contact (Fig.~\ref{figure:data_collection}), which we will refer to with $I_D \in \mathbb{R}^{H \times W}$. We use the depth images instead of RGB images as proposed by previous works~\cite{collins2023force} because the relationship between cantilever beam-like bodies such as soft robot fingers and contacting force is a well-defined problem in mechanics~\cite{liu2021effect}. Additionally, we want to capture MOE deformations without visual or depth distractions in the background. Previous works focused on collecting interaction datasets in visually diverse environments. For hair-care tasks, collecting sufficiently diverse task-relevant visual training data of MOE interacting with hair may be difficult. In our force estimation module we use Segment Anything Model 2 (SAM 2) vision foundation model~\cite{ravi2024sam} to generate a binary mask $M \in \{0,1\}^{H\times W}$. We prompt in an RGB image from the wrist-mounted RGBD camera with negative and positive prompt points based on the pixels corresponding to the MOE fingers as we show in Fig~\ref{figure:force_module}. We zero out the depth image outside of the observed pixels on the MOE fingers with the pixel-wise operation $I_D'(x,y) = I_D(x,y)M(x,y)$ and use the segmented depth image $I_D'$ as the visual observation for the force estimation module.

To infer MOE's applied forces, we use MOE actuators' current loads and the visual observation. Fig.~\ref{figure:deformed} shows MOE with different tendon tensions to reach various pose configurations. When MOE contacts a surface, even as delicate as a human user, the fingers deform. During deformation from contact, the increased tension on the tendons is observable from the servo actuators' increased current load (Fig.~\ref{figure:deformed}). The relationship between the tendon tension in an ideal condition without gear backlash and friction loss is $T= k_S q_i$, where $T$ is the tendon tension, $k_S$ is the system constant which considers the actuator characteristics and the pulley diameter, and $q_i$ is the induced actuator current load for actuator $i$ in MOE. For the two-finger MOE in this paper, we use four actuators to control eight tendons resulting in the MOE actuators' current load observation $\mathbf{q}\in \mathbb{R}^4$.

\begin{figure}

  \includegraphics[width=0.9\columnwidth]{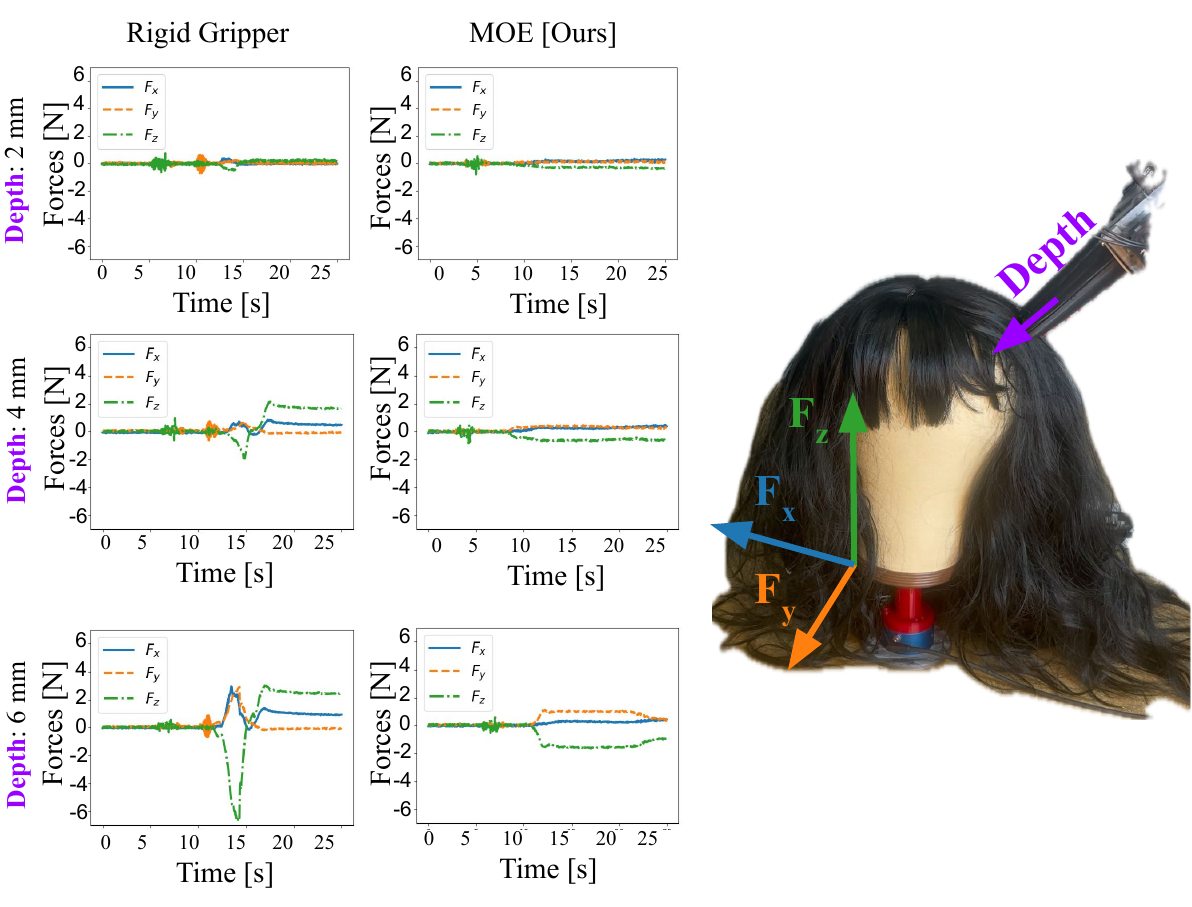}
  \caption{\textbf{Forces during hair grasping.} We carried out the experiments at three different depths into the hair. The depth measurements were from the point where the end-effector just made contact with the hair to account for different lengths. MOE exerts measurably less force and torque on the head. }
  \label{figure:task1results}
\end{figure}

\begin{table}
\caption{Hair Grasping Evaluation}
\setlength\tabcolsep{0pt} 
\begin{tabular*}{\columnwidth}{@{\extracolsep{\fill}} ll ccc}
\toprule
    End-effector & Depth [mm] & 
     \multicolumn{2}{c}{Performance Metrics} \\ 

     & &  Max Force [N] $\downarrow$& Grasped Hair [mm]  $\uparrow$\\
\midrule
     & 2.0 &  1.11 & 4.0 \\
    Rigid & 4.0 &  3.38 & 20.0 \\
     & 6.0 &  7.67 & 25.0 \\
    \cmidrule(){1-4}
     & 2.0 &  \bf 1.09 & \bf 5.0 \\
    MOE & 4.0 &  \bf 1.38 & \bf 18.7 \\
     & 6.0 &  \bf 1.98 & \bf 22.5 \\   
\bottomrule
\end{tabular*}
\scriptsize    
\label{tab:results} 
\end{table}

With the observations depth image $I_D'$, and the MOE actuator current load $\mathbf{q}$, the goal of the force estimation module is to learn the mapping $\hat{\mathbf{w}} =f(I_D, \mathbf{q})$, where $\hat{\mathbf{w}} \in \mathbb{R}^3$ is the estimated force vector. We use ResNet-18 image encoder~\cite{he2016deep} to encode $I_D'$ and an MLP to encode $\mathbf{q}$ each into 64-dimensional feature vectors that we concatenate into a 128-dimensional observation representation. Because of the design of MOE where the two fingers are aligned so that they can apply the most amount of force in the $z$-direction (Fig~\ref{figure:force_module}), the $z$-direction force component is notably larger than the other two force components. To address the magnitude imbalance in the dataset and encourage the model to correctly estimate the direction of the forces, we train the network with the weighted mean squared error loss 
\begin{equation}
    \mathcal{L}(\mathbf{w},\hat{\mathbf{w}})  = \|\mathbf{\lambda} \odot (\mathbf{w} - \hat{\mathbf{w}})\|_2^2,
\end{equation}
where $\mathbf{\lambda} \in \mathbb{R}^3 $ is the weighing vector for each direction components of the label and predicted forces $\mathbf{w},\hat{\mathbf{w}} \in \mathbb{R}^3 $ and $\odot$ denotes the Hadamard product. The force estimation module runs at 10.2\,Hz on the system's NVIDIA RTX 4090 GPU.

\subsection{Visual Perception Module}

Assistive robot systems must be able to perceive and track the user's pose to be effective and safe~\cite{jenamani2024feel}. In this work, we are primarily concerned with tracking the user's head pose. We employ the MediaPipe perception model~\cite{lugaresi2019mediapipe} to extract the visual key points in the user's face from the third-person RGBD camera's RGB image and get the corresponding 3D points, which allows us to get head tracking results at 12.5\,Hz. Fig.~\ref{fig:system_setup} shows the use of a visual marker on the tabletop to calibrate the coordinate frames. The user's 3D face key points are transformed into the robot frame for the subsequent hair manipulation tasks. For the finger combing task, we also use the MediaPipe model to extract key points on the participant's hand to record the desired trajectory for MOE to track. 

\begin{figure}
  \centering
    \includegraphics[width=0.8\linewidth]{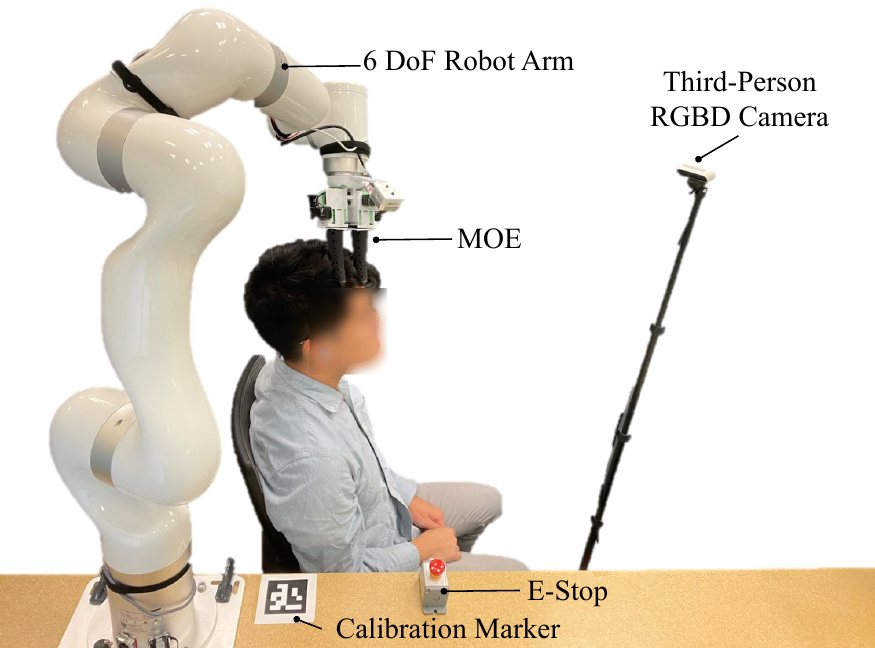}
    \caption{\textbf{The MOE-Hair hardware system setup}. MOE is attached to a 6 degrees of freedom (DoF) robot arm. A third-person RGBD camera tracks key points of the head and the participant's hand. A calibration marker is used to transform between camera and robot coordinate frames. }
    \label{fig:system_setup}
\end{figure}

\subsection{MOE-Hair System Integration}
\label{sec:system}
We integrate force estimation and the visual perception modules to build the MOE-Hair system to enable a force-reactive compliant soft robot hair manipulation (Fig.~\ref{fig:moehair_system}). MOE-Hair focuses on three hair manipulation and care skills: 1.~\emph{Head patting}, where MOE approaches the user's head from either the top or side to pat it; 2.~\emph{Finger combing}, where MOE follows a user-defined trajectory across the user's head; and 3.~\emph{Hair grasping}, where MOE approaches the user's head from either the top or side and grasps hair from near the scalp.

To perform these tasks effectively, we use the visual perception module to perceive the user's head pose and 3D key points in the robot frame. We use the approximate head pose to approach the user's head with MOE. Once in contact, we use the force estimation module in a force-feedback loop and allow the system to apply consistent force on the scalp, even with imperfect visual input. The system adjusts the robot's poses in real-time to adapt to the user's head movements.

\begin{figure}
  \centering
    \includegraphics[width=1.0\linewidth]{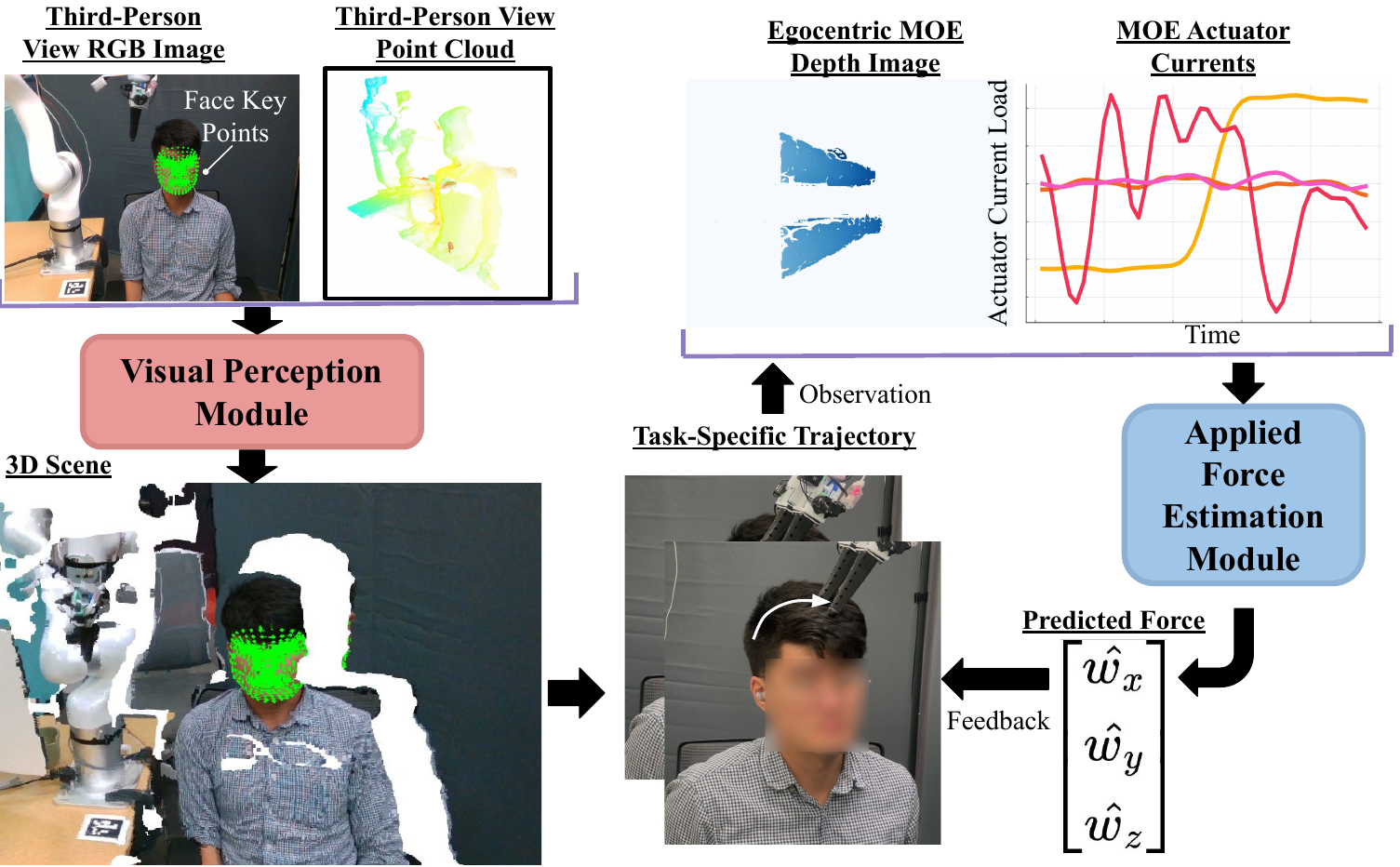}
    \caption{\textbf{The MOE-Hair system integration}. We use a third-person RGBD camera to track the user's visual face key points. The visual perception module maps the visual key points to the 3D point cloud and generates a task-specific trajectory. As MOE follows the trajectory, the system predicts the applied force on the head using the segmented depth images of the MOE fingers and actuator current load. The system adjusts the trajectories in real time to maintain comfort and contact based on the predicted contact forces. }
    \label{fig:moehair_system}
\end{figure}

\section{System Results}
We present evaluation results for the components of the proposed MOE-Hair system. Specifically, we demonstrate that the proposed MOE soft robot end-effector can interact with a surface while applying lower contact forces compared to a rigid end-effector. Furthermore, we show that we achieve improved applied force estimation using both observed depth and actuator current loads, compared to using only depth or actuator load. 

\subsection{MOE Interaction Force Evaluation}
We use a mannequin head with an attached synthetic hair wig to simulate a human user. The base of the mannequin head is rigidly mounted to a 6-axis force sensor to measure the applied forces on the head. We compared the applied forces with open-loop experiments, where a rigid parallel jaw gripper (FE Gripper, Franka Robotics) and proposed MOE moved to a specified depth (2.0\,mm, 4.0\,mm, 6.0\,mm) into the hair and grasped. The depths are measured with respect to the position where the robot is barely making contact with the hair to account for different end-effectors lengths. After the grippers grasped the hair, the robot hands moved up to lift the grasped bundle of hair. We then measured the minimum packing perimeter of the bundle of the grasped hair to assess task effectiveness. Fig.~\ref{figure:task1results} shows the forces experienced by the force-sensorized mannequin head during the grasping task.


\begin{figure}[t!]
\centering
\includegraphics[width=0.9\columnwidth]{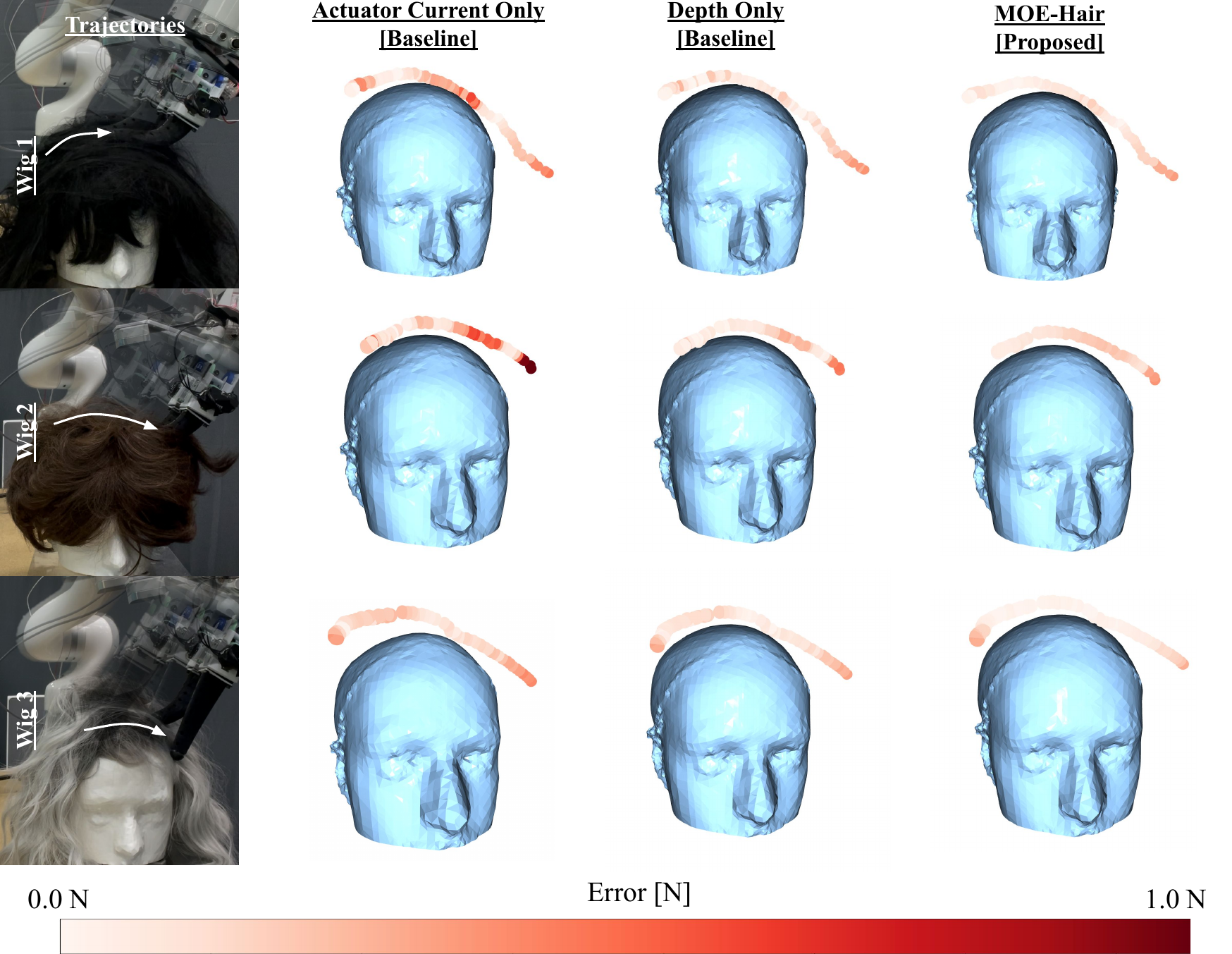}
  \caption{\textbf{Visual comparison of the applied force estimation errors}. We compare between the ground-truth forces experienced by the mannequin head with different wigs, to the predicted forces from the baselines and the proposed approach.}
  \label{figure:force_estimation_results}
\end{figure}

\begin{table}
\caption{Force Estimation Comparisons}
\setlength\tabcolsep{0pt} 
\begin{tabular*}{\columnwidth}{@{\extracolsep{\fill}} l l c}
\toprule
    \textbf{Wig} & \textbf{Method} & \textbf{RMSE (N)} $\downarrow$ \\ 
\midrule
    \multirow{3}{*}{Wig 1} & Actuator Load Only [Baseline] & 0.286 \\ 
                           &  Depth Only [Baseline]  &  0.143 \\ 
                           & MOE-Hair [Proposed] & \textbf{0.114} \\ 
                   
    \cmidrule(){1-3}
    \multirow{3}{*}{Wig 2} & Actuator Load Only [Baseline]&  0.318 \\ 
                            &  Depth Only [Baseline] &  0.185 \\ 
                            & MOE-Hair [Proposed] &  \textbf{0.162}  \\
    \cmidrule(){1-3}
    \multirow{3}{*}{Wig 3} & Actuator Load Only [Baseline]& 0.296 \\ 
                           &  Depth Only [Baseline] &  0.154 \\ 
                           & MOE-Hair [Proposed] & \textbf{0.133} \\ 

\bottomrule
\end{tabular*}
\scriptsize
\label{tab:mse_results}
\end{table}

Lower forces experienced by the mannequin head could indicate reduced discomfort if applied to a human subject. Concurrently, a hair care robot will need to be able to grasp hair that may be close to the scalp, which will likely result in higher forces experienced by the mannequin head. Then, we note that an ideal hair care robot must be able to grasp hair effectively while also applying minimal force on the head. Table~\ref{tab:results} reports the maximum force experienced by the head at varying depths and the amount of hair grasped. 

At 6.0\,mm depth, the rigid end-effector exerts 7.67\,N of force on the mannequin head, while MOE applied 1.98\,N of force. This constitutes a 74.1\,\% reduction in the maximum force applied to the head. Meanwhile, on the grasped hair metric, MOE grasped approximately 10\,\% less hair. A potential explanation of this marginal decrease in the amount of hair grasped is that the compliance of MOE allowed some of the grasped hair to be pried away as the end-effector moved away. From these results, we design MOE-Hair's force-feedback controller to maintain an estimated applied force of 2.0\,N to optimize for task effectiveness.  

\subsection{Applied Force Estimation Evaluation}
We evaluated the force estimation module with the force-sensorized mannequin head and the finger combing task. We chose the finger combing task because it has the longest contact duration of the three studied tasks. We compare the force estimation module to two baselines: (1)~using only actuator load, inspired by previous works predicting contact based on tendon tension changes~\cite{liu2021effect,liu2021influence}; and (2)~using only depth images, inspired by~\cite{grady2022visual}.

For three wigs with different lengths and color, we have MOE follow a finger combing trajectory and record the depth image and actuator current load observation along with the synchronized ground-truth force readings from the sensor attached to the mannequin head. We visualize the root mean squared error (RMSE) over the trajectory in Fig.~\ref{figure:force_estimation_results} and report them in Table~\ref{tab:mse_results}. In all three cases, the proposed MOE-Hair applied force estimation module outperforms both baselines, improving the actuator load-only and depth image-only baselines by up to 60.1\,\% and 20.3\,\%, respectively.

\section{User Study Evaluation}

\begin{figure*}
  \centering
    \includegraphics[width=0.85\linewidth]{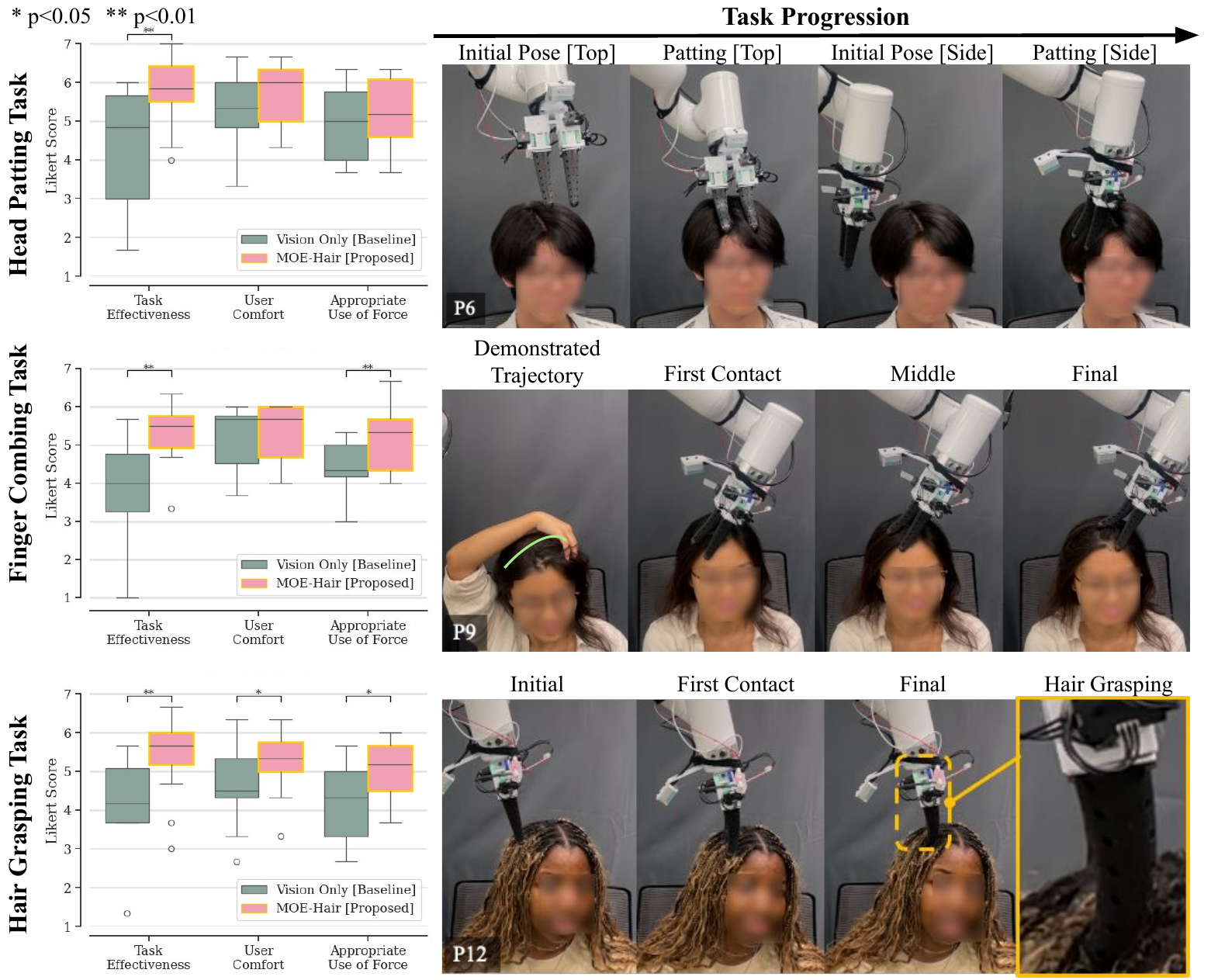}
    \caption{\textbf{User study results}. We evaluate on perceived task effectiveness, user comfort, and appropriate use of force. Top row: results for the head patting task where MOE pats the user's head from the top and the side. Middle row: results for the finger combing task where MOE follows the user's demonstrated trajectory to comb the hair. Bottom row: results for the hair grasping task where MOE approaches the user's head and grasps with its dexterous soft fingers.  }
    \label{fig:user_results}
\end{figure*}

We study three tasks to evaluate the MOE system: head patting, finger combing, and hair grasping. For each of these tasks, we compared the proposed MOE-Hair system that uses force feedback, to vision-only with no force feedback. To evaluate our system, we use three constructs proposed by previous works as important components of hair-care systems~\cite{dennler_design_2021,hirose2012development}: \textbf{task effectiveness}, \textbf{user comfort}, and \textbf{appropriate use of force}. We performed a 3 (task) by 2 (system) within-subjects user study to evaluate the system.

\subsection{Hypotheses}
Given the importance of force feedback in assistive domains~\cite{jenamani2024feel}, we expect the proposed MOE-Hair system will be rated more favorably by participants compared to the vision-only baseline. We therefore developed three hypotheses:\\
\textbf{H1: } \textit{MOE-Hair will perform the tasks more effectively by using force feedback and thus participants will rate MOE-Hair higher on \textbf{task effectiveness} than the vision-only baseline.}

\noindent\textbf{H2: } \textit{MOE-Hair will not exert forces that are too high due to its force feedback so participants will rate MOE-Hair higher in \textbf{user comfort} compared to the vision-only baseline.}

\noindent\textbf{H3: } \textit{MOE-Hair will consistently apply appropriate contact forces to the head and thus participants will rate MOE-Hair higher in \textbf{appropriate use of force} compared to the vision-only baseline.}

\subsection{Participants}
We recruited 12 participants (ages 20--27, hair lengths 25--508\,mm) without visual or mobility limitations as this is the first work with user studies on robot haircare. The participants represented diverse hair types based on an accepted classification~\cite{de2007shape}.


\subsection{Procedure}
 Before interacting with the robot, we asked the participants to complete a demographic and pre-study questionnaire. We asked the users to sit in a chair facing the third-person view camera for the duration of the study, except to answer the questionnaires after each trial. For each task, users experienced both force feedback conditions---MOE-Hair and the vision-only baseline---in a randomized and counterbalanced order. After each interaction with the system, the participants were asked to complete 7-point 3-item Likert scale for each of the three constructs we measured: task effectiveness, user comfort, and appropriate use of force for the task. After the user experienced both force feedback conditions, the participants were asked to complete a questionnaire regarding their preferences on which system performed better for task effectiveness, user comfort, and appropriate use of force for the task. Based on literature~\cite{holmes2010hair} and common hair care practices, we selected three tasks with increasing levels of physical invasiveness, arranged in the following order:

\textbf{Head Patting.} For each system, MOE approached the top of the participant's head first. Then both systems patted the head three times. The systems reset to the home positions and approached the participants' heads from their right-hand side to pat the head three times. The goal of the systems are to compress the hair to the scalp consistently across the three patting opportunities as opposed to superficial contact with the hair. The average task completion time was 15 seconds.

\textbf{Finger Combing.} We asked the participants to first demonstrate to the systems their desired trajectories for the robot to follow by moving their hand across the head and making contact with their scalps. The visual tracking module recorded the participant's 3D hand trajectory. Then, systems tracked the demonstrated trajectories. The average task completion time was 38 seconds.

\textbf{Hair Grasping.} Similar to the head-patting task, MOE approaches the top of the participant's head first, and each of the systems is tasked with compressing the hair to the scalp. Then, the MOE fingers are actuated to grasp the hair and hold it in the hand for three seconds. Afterward, MOE ungrasps the hair and returns to the home position. The average task completion time was 24 seconds. After all of the tasks were completed, we conducted a semi-structured interview with the participants. The entire study took approximately 40 minutes.

\subsection{Responses}
Fig.~\ref{fig:user_results} shows the participant responses for task effectiveness, user comfort, and the system's appropriate use of force in each of the tasks for both force feedback conditions. Additionally, we report participant responses for their preferred force feedback condition in Fig.~\ref{fig:pref}, where higher Likert scale rating corresponds to a preference for MOE-Hair. We conducted non-parametric Wilcoxon signed-rank tests~\cite{wilcoxon1992individual} for paired user responses based on the two systems performing the same tasks from Fig.~\ref{fig:user_results}. We also evaluated if the participants had a significant preference for a system from the neutral value of 4 with the responses in Fig.~\ref{fig:pref}.

\noindent\textbf{Task Effectiveness.} Across all three tasks, participants rated the task effectiveness of MOE-Hair as higher than the vision-only baseline (all $p{<}.01)$. Additionally, we found a statistically significant preference for MOE-Hair for all three tasks for the comparison question of system effectiveness. The finding supports \textbf{H1} that users will find MOE-Hair to be more effective for the tasks compared to the vision-only system.

\noindent\textbf{User Comfort.} The participants reported being comfortable with both systems in all of the tasks, generally agreeing that the system put them at ease and that they trusted the system to be safe. However, we only found significant differences in the hair-grasping task, where participants rated MOE-Hair as more comfortable than the vision-only baseline ($p=.02$). We found no statistically significant preference for either force feedback condition on user comfort, although the participants tended to be more comfortable with the MOE-Hair system. This partially supports \textbf{H2} with the caveat that the biggest differences in comfort are for more involved tasks. The vision-only system sometimes failed to make consistent scalp contact, which some participants found similarly comfortable due to the lack of direct contact.

\noindent\textbf{Appropriate Use of Force.} Notably, the comparison of systems' perceived appropriate use of force in the finger combing task yielded statistically significant differences in favor of MOE-Hair. This finding supports \textbf{H3} for the finger combing task, which was the task with the longest duration of continued contact between the MOE fingers and the participant's head. Although the response averages favored MOE-Hair for the appropriate use of force for the head-patting and hair-grasping tasks, the differences were not statistically significant. However, we found a statistically significant preference of the participants in favor of MOE-Hair on the system's appropriate use of force for all three tasks, supporting \textbf{H3}.

\begin{figure}[t!]
\includegraphics[width=1\linewidth]{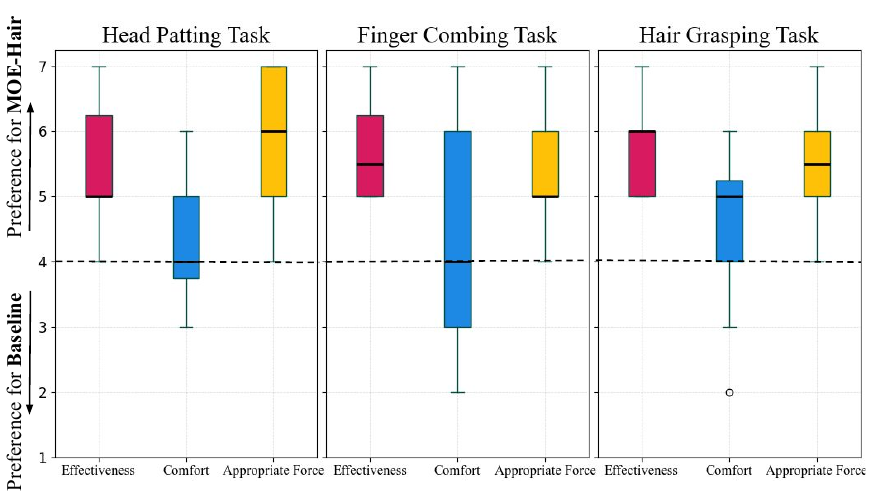}
  \caption{\textbf{User preferences of the methods} for the three tasks. Lower Likert scale score indicates preference for the baseline system with vision feedback only and higher score indicates preferences for the proposed MOE-Hair system. }
  \label{fig:pref}
  \vspace{-0.5 cm}
\end{figure}

\subsection{Qualitative Results}
Based on the participants' responses to open-ended interview questions, three themes emerged, relating to the evaluation of the presented robot system. 

\textbf{Effectiveness.} All 12 of the participants responded positively to the system's ability to perform the three skills outlined in this work. The participants indicated that MOE effectively performed the tasks and P6 noted, ``It was working reliably.'' P12 noted, ``I didn't think it was going to work on my braids but it actually did well.'' Similarly, P2 remarked ``I've been told my hair is difficult because it's so thick'' and commented that MOE-Hair performed effectively, noting ``It did pretty well.'' The responses underscored the system's ability to perform well with varying hair characteristics and conditions.

Negative responses to the system's effectiveness are related primarily to the participant's uncertainty about the capabilities of the system. 3 of the participants responded that they wished the system could perform more tasks such as hair cutting and styling, where P11 commented ``it would've been nice if it could cut my hair,'' so that ``I don't have to talk to [human hairstylists].''

\textbf{User tactile sensation.} All 12 of the participants noted that they generally felt safe. For example, P7 noted that they ``felt really comfortable.'' 5 participants commented that MOE's contact sensation was pleasant. P1 liked the sensation when MOE made contact, remarking ``It was like a head massage.''  P2 noted that MOE fingers felt ``really similar to [human] fingers'' and that ``I kind of forgot it was a robot arm for a minute.'' P11 stated that MOE ``made me relax.''

Negative responses highlighted some participants' hesitancy with an autonomous robot system touching their heads. P6 noted that MOE-Hair ``felt like it's kind of pushing my head.'' P12 noted that they were at first uncertain if the system was ``safe or not'' with the first task but stated that they became ``more comfortable'' with the latter tasks, suggesting the process of MOE-Hair system building trust with its users.

\textbf{Appearance of MOE.} 8 participants made generally positive comments about MOE's appearance. As P4 noted, ``I felt safe because it looked rubbery and soft.'' 3 participants noted that they liked that MOE did not look anthropomorphic and P5 stated MOE ``just looks like it's its own thing and it doesn't give me an uncanny valley feeling.'' 

2 participants noted that the wires of the actuators contributed to their unease with the system. For example, P1 noted MOE ``seemed like a prototype and doesn't seem like a product'' because of the visible wires. P7 noted MOE ``doesn't look terrible'' but that ``the cables could be cleaned up.'' P12 made comments on MOE's color and shape commenting ``it doesn't look friendly.'' Diverging comments on the appearance and visual behavior of the robot could indicate a need to incorporate design insights from literature~\cite{dennler2023design} to be more likely to be perceived as being friendly by the users.

\section{Discussion and Limitations}
In this work, we introduce a soft robot manipulator for hair manipulation and care tasks that we call MOE. The results suggest that in comparison to its mechanically rigid counterparts, MOE is safer in close contact with a head and that observing deformations of MOE can provide us with useful information about the applied forces on human users. The experimental results with the mannequin testbed showed that soft robots such as MOE could be effectively exploited in hair care tasks. Human user study highlighted that the proposed MOE-Hair system with the contact force estimation module was perceived favorably in its effectiveness for the tasks, comfort, and ability to use appropriate forces. Qualitative analysis of the participants' responses to open-ended questions revealed that some also perceived the tactile sensation of MOE to be pleasant and relaxing, which seems to offer promising extension in applying MOE to other contact-rich pHRI tasks.

A limitation of the study was in the scope of the presented tasks. Building on the results of this study indicating broadly high perceived comfort with the MOE-Hair system, future works will introduce expanding capabilities for the system to incorporate user-defined goals to assist in other hair care tasks. 

Another limitation is in the system's behavior generation. The focus of the MOE-Hair system development was on effective and safe contact using the compliance of the MOE soft fingers and force feedback with the proposed applied force estimation module. However, the characteristics of movement and personalization of robot trajectories could be important factors for users viewing the system favorably~\cite{dennler_design_2021}. To this end, we can improve MOE-Hair system to incorporate strategies for planning trajectories with human preferences. Future work will also explore customizing force thresholds and approach speeds for elderly and mobility-impaired users to maximize safety and comfort.

\section*{Acknowledgment}
This work is supported by the NSF GRFP (Grant No. DGE2140739) and MOTIE, Korea (Grant No. 20018112).

\balance
\bibliographystyle{ieeetr}
\bibliography{ref}

\newpage
\nobalance
\appendices

\section{User Study Participants}
\begin{table}[htbp]
\centering
\begin{tabular}{|c|c|c|c|c|c|}
\hline
\textbf{ID} & \textbf{Age} & \textbf{Gender} & \textbf{Ethnicity} & \textbf{Hair Length [mm]} & \textbf{NARS Scale [7-pt]} \\ \hline
P1 & 26  & Man   & Asian  & 75   & 4.6  \\ \hline
P2 & 22 & Man  & Latino & 400 & 2.2 \\ \hline
P3 & 24 & Man & Asian  & 110  & 2.6  \\ \hline
P4 & 25 & Man & Latino & 50  & 1.4 \\ \hline
P5 & 27  & Man & S Asian & 100 & 3.2 \\ \hline
P6 & 20 & Man  & Asian  & 80 & 4.2 \\ \hline%
P7 & 26 & Woman  & Asian  & 381  & 3.2 \\ \hline
P8 & 27 & Man  & White & 45 & 3.0 \\ \hline
P9 &  24  & Woman & Asian  & 400  & 3.2 \\ \hline%
P10 & 27&  Man & S Asian & 50 & 3.2 \\ \hline
P11 & 25 & Man & Asian & 65 & 3.6 \\ \hline
P12 & 25 & Woman & Black  & 508  & 4.6  \\ \hline
\end{tabular}
\caption{Demographic Data with NARS Scores}
\label{tab:demographic_data}
\end{table}

\section{Additional Force Results}

\begin{figure}[h!]
    \centering
    \includegraphics[width=1.2 \columnwidth]{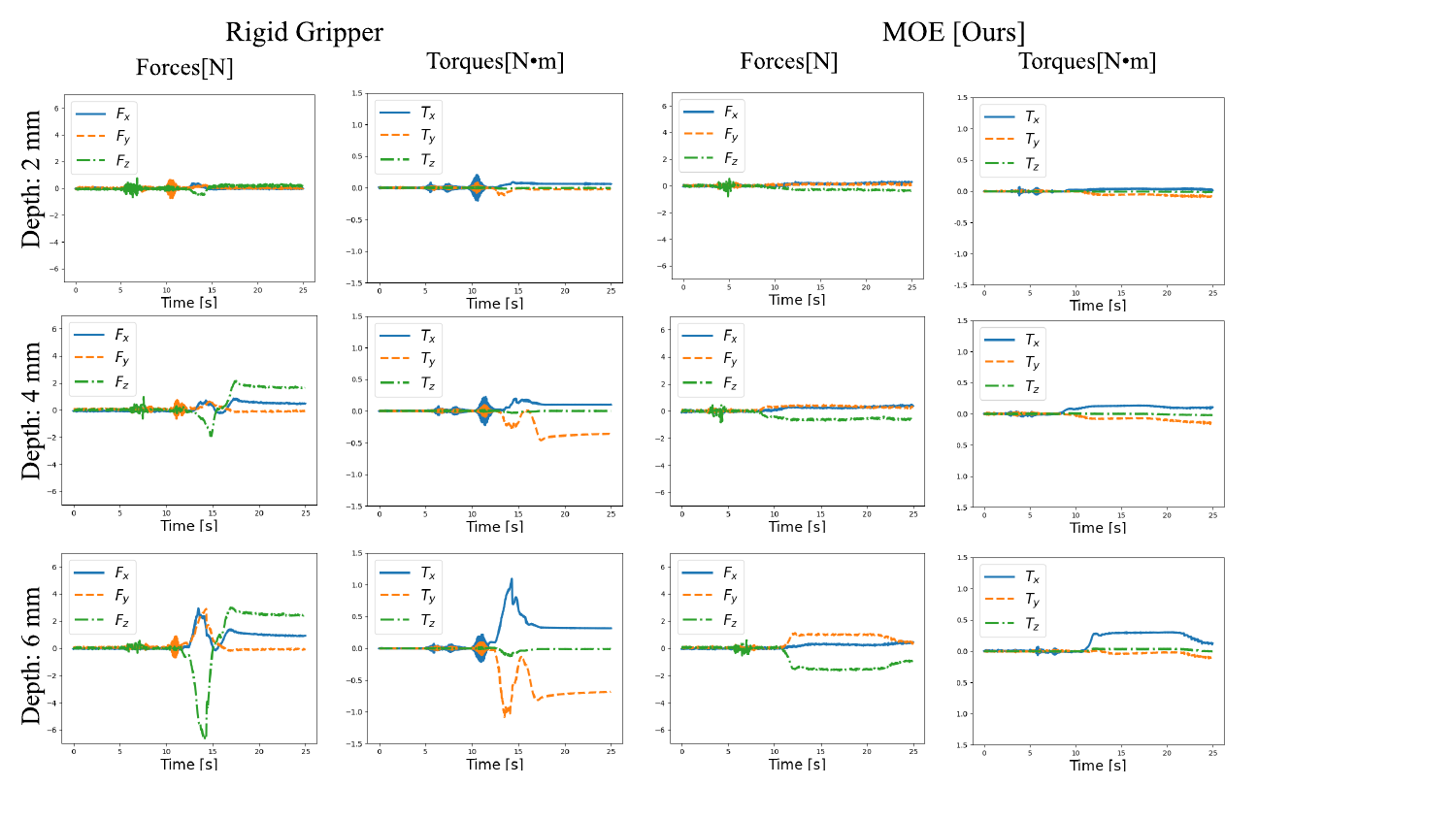}
    \caption{\textbf{Full set of force and torque readings} for the comparison between a rigid gripper and proposed MOE.}
\end{figure}

\begin{figure}[h!]
    \centering
    \hspace{-1.0 cm}\includegraphics[width=1.1 \columnwidth]{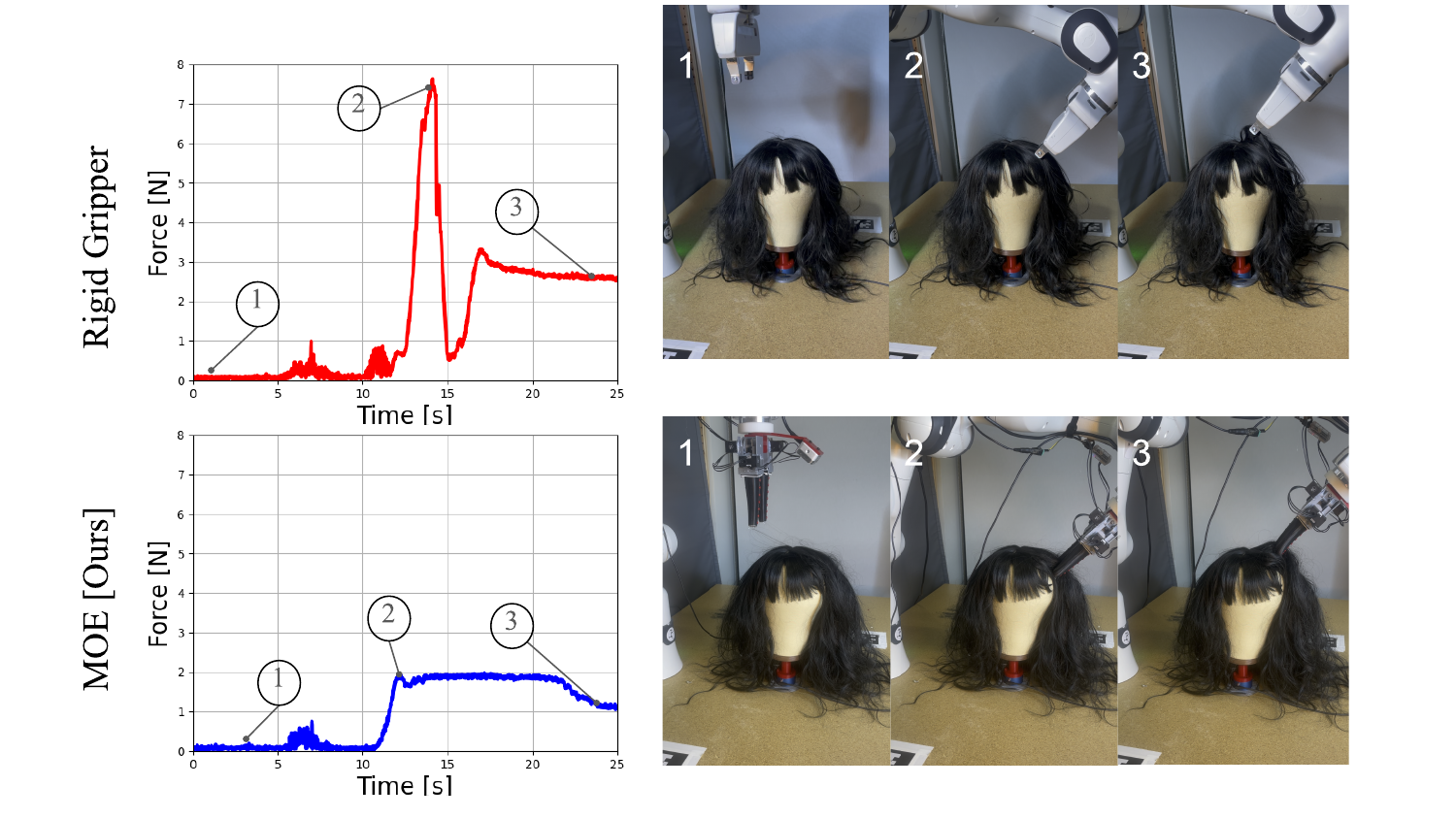}
    \caption{\textbf{Hair grasping task progression on a sensorized mannequin head. }}
\end{figure}
\vfill

\section{Questionnaire}

\begin{figure}[h!]
    \centering
    \includegraphics[width=0.8 \columnwidth]{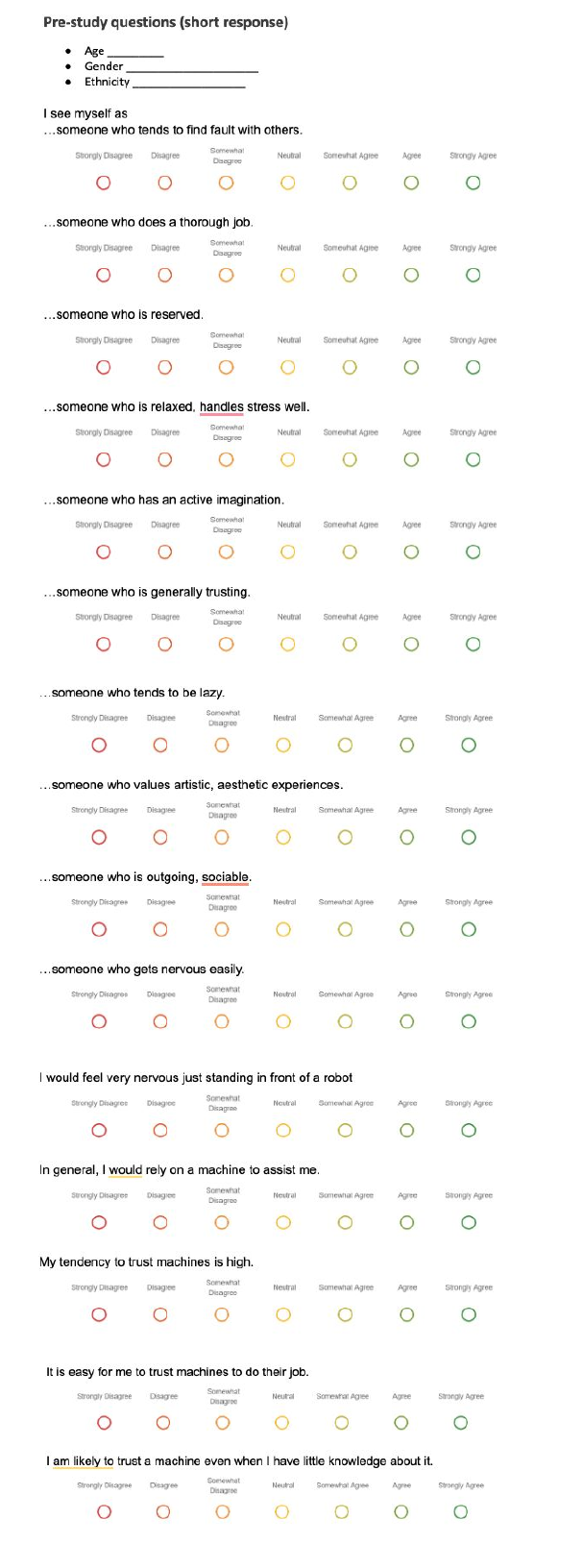}
    \caption{\textbf{Pre-Study Questionnaire.} }
\end{figure}

\begin{figure}[h!]
    \centering
    \includegraphics[width=0.8 \columnwidth]{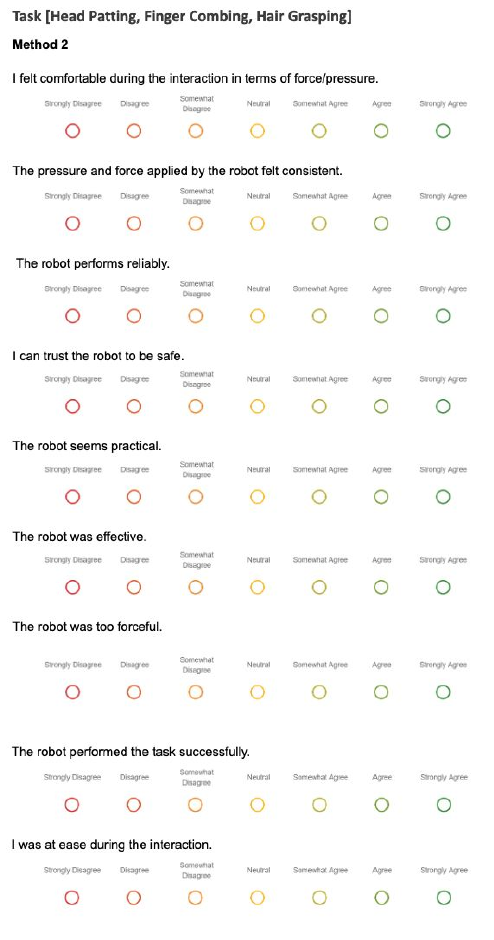}
    \caption{\textbf{Questionnaire given to participants after a trial with a system for a task.} }
\end{figure}

\begin{figure}[h!]
    \centering
    \includegraphics[width=1.0 \columnwidth]{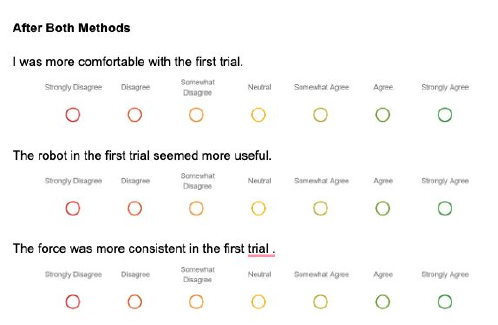}
    \caption{\textbf{Questionnaire given to participants after trials with both systems for a task.} }
\end{figure}

\newpage
\section{Institutional Review}
The Institutional Review Board approved the protocol for the trials and the study (STUDY2023\_00000502).
\end{document}